\documentclass[journal]{IEEEtran}
\usepackage[utf8]{inputenc}
\usepackage{amsmath,amssymb,amsfonts}
\usepackage{algorithm} 
\usepackage{epstopdf}
\usepackage{subcaption} 
\usepackage{caption}    
\usepackage{textcomp}
\usepackage{bm}
\usepackage{balance}
\usepackage{amsmath,amsfonts}
\usepackage{algorithmic}
\usepackage{array}
\usepackage{textcomp}
\usepackage{stfloats}
\usepackage{url}
\usepackage{verbatim}
\usepackage{graphicx}
\usepackage{booktabs}
\usepackage{balance}
\usepackage{multirow}
\usepackage{diagbox}
\def\BibTeX{{\rm B\kern-.05em{\sc i\kern-.025em b}\kern-.08em
		T\kern-.1667em\lower.7ex\hbox{E}\kern-.125emX}}
\allowdisplaybreaks

\vspace{-10.pt}
\begin{document}
	\title{\huge Performance Comparison of Aerial RIS and STAR-RIS in 3D Wireless Environments}
	\author{Dongdong~Yang, Bin~Li, Jiguang~He,~\IEEEmembership{Senior Member,~IEEE}
		\thanks{D. Yang is with Nanjing University of Information Science and Technology, Nanjing 210044, China, and also with Great Bay University, Dongguan 523000, China (e-mail: 202312200024@nuist.edu.cn).}
		\thanks{B. Li is with Nanjing University of Information Science and Technology, Nanjing 210044, China (e-mail: bin.li@nuist.edu.cn).}
		\thanks{J. He is with Great Bay University, Dongguan 523000, China (e-mail: jiguang.he@gbu.edu.cn).}
	}
	
	\setlength{\parskip}{0pt} 
	\maketitle
	
	\vspace{-15.pt}
	\begin{abstract}
	Reconfigurable intelligent surface (RIS) and simultaneously transmitting and reflecting RIS (STAR-RIS) have emerged as key enablers for enhancing wireless coverage and capacity in next-generation networks. When mounted on unmanned aerial vehicles (UAVs), they benefit from flexible deployment and improved line-of-sight conditions. Despite their promising potential, a comprehensive performance comparison between aerial RIS and STAR-RIS architectures has not been thoroughly investigated. This letter presents a detailed performance comparison between aerial RIS and STAR-RIS in three-dimensional wireless environments. Accurate channel models incorporating directional radiation patterns are established, and the influence of deployment altitude and orientation is thoroughly examined. To optimize the system sum-rate, we formulate joint optimization problems for both architectures and propose an efficient solution based on the weighted minimum mean square error and block coordinate descent algorithms. Simulation results reveal that STAR-RIS outperforms RIS in low-altitude scenarios due to its full-space coverage capability, whereas RIS delivers better performance near the base station at higher altitudes. The findings provide practical insights for the deployment of aerial intelligent surfaces in future 6G communication systems.
	\end{abstract}
	
	\section{Introduction}
	 The advent of sixth-generation (6G) wireless networks demands unprecedented capabilities in terms of ubiquitous coverage, ultra-high data rates, and dynamic adaptability to complex environments, such as smart cities and integrated air-ground-space networks. Traditional base station (BS)-centric architectures face inherent limitations, including high deployment costs, energy inefficiency, and inflexibility in practical scenarios \cite{9424177}. To address these challenges, reconfigurable intelligent surface (RIS) has emerged as an essential technology that enhances wireless propagation environments through intelligent electromagnetic wave manipulation \cite{10480441}. While conventional RIS operates solely in reflection mode, the recently proposed simultaneously transmitting and reflecting RIS (STAR-RIS) breaks this limitation, enabling full-space coverage via dual signal transmission and reflection capabilities \cite{9437234,9856598}. However, conventional terrestrial scheme is constrained by its fixed deployment, limiting service area to static coverage regions \cite{10345491}. The limitation can be mitigated by integrating RIS and STAR-RIS with unmanned aerial vehicle (UAV), renowned for its high line-of-sight (LoS) probability and three-dimensional (3D) maneuverability \cite{8579209}. The resultant aerial RIS and STAR-RIS architectures synergistically integrate the complementary benefits of both technologies, establishing itself as a promising solution for next-generation networks \cite{9749767}.
	 
	 Existing studies have predominantly focused on optimizing aerial RIS or STAR-RIS systems. For example, Liu \textit{et al.} \cite{liu2022throughput} jointly optimized aerial RIS trajectory and dynamic power allocation to maximize average downlink throughput in time-slotted transmissions. 
	 Considering the influence of adversarial eavesdropping and malicious jamming, Yang \textit{et al.}~\cite{10345491} proposed a reliable and secure communication scheme assisted by aerial RIS, and jointly designed the transmission beamforming, artificial noise, aerial RIS's placement and phase shifts to maximize the achievable worst-case secrecy rate.
	 Furthermore, Aung \textit{et al.} \cite{10458888} introduced the aerial STAR-RIS into the mobile edge computing system and utilized a deep reinforcement learning-based approach to design the UAV trajectory, STAR-RIS configurations, and task offloading strategies. Xiao \textit{et al.} \cite{10850614} proposed a STAR-RIS-enhanced UAV-assisted mobile edge computing scheme with bi-directional task offloading, jointly optimizing resources, scheduling, beamforming, and trajectory to maximize energy efficiency under quality of service constraints.
	 
	 However, the performance trade-offs between aerial RIS and STAR-RIS architectures remain underexplored. Therefore, in this paper, we first perform accurate channel modeling based on a 3D radiation pattern to compare a horizontally deployed RIS and a vertically deployed STAR-RIS. Furthermore, optimization problems are formulated in both scenarios aiming to maximize the sum transmission rate, and a solution framework based on the block coordinate descent (BCD) and weighted minimum mean square error (WMMSE) methods is proposed. Finally, simulation results demonstrate that each architecture exhibits respective performance advantages: the STAR-RIS performs better at low altitudes, while the RIS achieves superior performance at higher altitudes near the BS.
	 
	 \section{System model}
	\subsection{Channel Model}
	In this paper, we consider an $M$-antenna BS, an $N$-element aerial RIS and STAR-RIS, and $K$ single-antenna users, denoted by ${\cal K} = \left\{ {1, \ldots ,k, \ldots ,K} \right\}$. 
	To ensure a fair comparison, the STAR-RIS is supposed to function in the energy splitting mode, wherein each element is capable of concurrently transmitting and reflecting incident signals. Furthermore, we assume that users in ${\cal K}_t$ are positioned on the transmission half-space, and  the users in ${\cal K}_r$ are positioned on the reflection half-space. The proposed aerial RIS and STAR-RIS architectures are illustrated in Fig. 1, where the RIS is horizontally deployed, and the STAR-RIS is vertically deployed, achieving full-space coverage through its dual capability of simultaneous signal transmission and reflection. 
	
	In practice, the aerial RIS and STAR-RIS are deployed to enhance channel quality by establishing LoS links. However, non-line-of-sight (NLoS) propagation still plays a significant role, particularly in environments with substantial scattering. Therefore, the channel  between the BS and RIS/STAR-RIS, as well as between the RIS/STAR-RIS and user $k$ are modeled as Rician fading channels, given by
	\begin{equation}
		{{\bf{H}}_{B,R}} = \sqrt {{\rho _{B,R}}} \left( {\sqrt {\frac{{{K_1}}}{{1 + {K_1}}}} {{{\bar{\bf H}}}_{B,R}} + \sqrt {\frac{1}{{1 + {K_1}}}} {{{ \tilde{\bf H}}}_{B,R}}} \right),
	\end{equation}
	\begin{equation}
		{{\bf{h}}_{R,k}} = \sqrt {{\rho _{R,k}}} \left( {\sqrt {\frac{{{K_2}}}{{1 + {K_2}}}} {{{\bar{\bf h}}}_{R,k}} + \sqrt {\frac{1}{{1 + {K_2}}}} {{{\tilde{\bf h}}}_{R,k}}} \right),
	\end{equation}
	where ${\bar {\bf{H}} _{B,R}} \in {\mathbb{C}^{N\times M}}$ and ${\bar {\bf{h}} _{R,k}} \in {\mathbb{C}^{N \times 1}}$ denote the LoS components, ${ \tilde{\bf H}_{B,R}}\in {\mathbb{C}^{N\times M}}$ and ${\tilde{\bf h}_{R,k}}\in {\mathbb{C}^{N \times 1}}$ represent the NLoS components, $K_1$ and $K_2$ are the Rician factors. Finally, ${\rho _{B,R}}$ and ${\rho _{R,k}}$ denote the large scale path-loss. According to \cite{9722711}, ${\rho _{B,R}}$ and ${\rho _{R,k}}$ are given by
	\begin{equation}
		{\rho _{B,R}} = \frac{{{\rho _0}}}{{d_{B,R}^2}}{D_B}{D_R}{\left| {{F_B}\left( {\theta _{B,R}^{{\rm{AoD}}},\varphi _{B,R}^{{\rm{AoD}}}} \right)} \right|^2}{\left| {{F_k}\left( {\theta _{B,R}^{{\rm{AoA}}},\varphi _{B,R}^{{\rm{AoA}}}} \right)} \right|^2},
	\end{equation}
	\begin{equation}
		{\rho _{R,k}} = \frac{{{\rho _0}}}{{d_{R,k}^2}}{D_R}{D_k}{\left| {{F_R}\left( {\theta _{R,k}^{{\rm{AoD}}},\varphi _{R,k}^{{\rm{AoD}}}} \right)} \right|^2}{\left| {{F_k}\left( {\theta _{R,k}^{{\rm{AoA}}},\varphi _{R,k}^{{\rm{AoA}}}} \right)} \right|^2},
	\end{equation}
	where $\rho_0$ is the path-loss factor at the reference distance, $d_{B,R}$ and $d_{R,k}$ denote the distance between the BS and RIS/STAR-RIS, as well as that between the RIS/STAR-RIS and user $k$, $D_B$, $D_R$, and $D_k$ are the maximum directivities of the BS, RIS/STAR-RIS, and user $k$. Moreover, $F_B({\theta, \varphi})$, $F_R({\theta, \varphi})$, and $F_k({\theta, \varphi})$ are the normalized radiation pattern of them. Specifically, $\theta^{\mathrm{AoD}}_{B,R}$ and $\varphi^{\mathrm{AoD}}_{B,R}$ denote the azimuth and elevation angles of departure (AoD) from the BS to RIS/STAR-RIS, $\theta^{\mathrm{AoA}}_{B,R}$ and $\varphi^{\mathrm{AoA}}_{B,R}$ denote the azimuth and elevation angles of arrival (AoA) from the BS to RIS/STAR-RIS. Similarly, $\theta^{\mathrm{AoD}}_{R,k}$/$\varphi^{\mathrm{AoD}}_{R,k}$ and $\theta^{\mathrm{AoA}}_{R,k}$/$\varphi^{\mathrm{AoA}}_{R,k}$ define the AoDs and AoAs from the RIS/STAR-RIS to user $k$ in both azimuth and elevation.
	\begin{figure}[t]
		\centering
		\includegraphics[width=1\columnwidth]{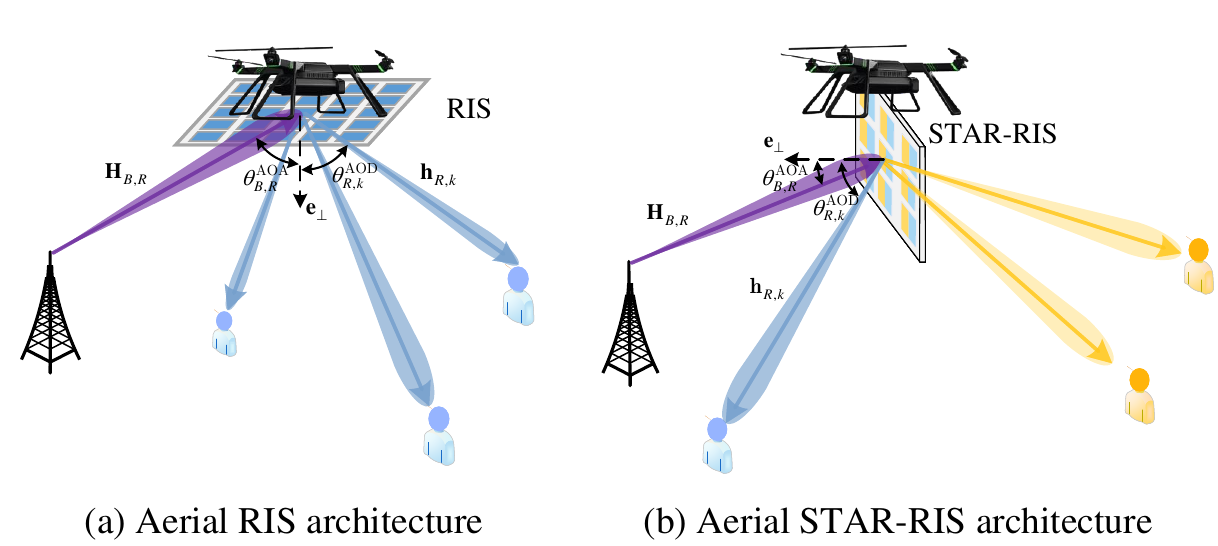}
		\caption{Aerial RIS and STAR-RIS architectures.} 
	\label{fig:model}
	\end{figure}
	
	For the antenna whose major lobe is oriented along the $z$-axis as illustrated in Fig. 1 (a), the radiation intensity is expressed as $U(\theta,\varphi)$, with $\theta$ and $\varphi$ being the elevation and azimuth AoDs from the antenna. Then, the general normalized power patterns can be modeled as \cite{balanis2016antenna}
	\begin{equation}
		U\left( {\theta ,\varphi } \right) = \left\{ {\begin{array}{*{20}{ll}}
				{{{\cos }^q}(\theta ),}&{\theta  \in \left[ {0,\frac{\pi }{2}} \right],\varphi  \in \left[ {0,2\pi } \right],}\\
				{0,}&{{\rm{otherwise,}}}
		\end{array}} \right.
	\end{equation}
	where the exponent $q$ characterizes the directivity. 
	
	The antenna directivity is formally defined as the ratio between the radiated power in a specific direction and the mean radiation intensity over the entire sphere, given by 
	\begin{equation}
		D = \frac{4\pi}{\iint\limits_{\Omega} U(\theta, \varphi) \, d\Omega} 
		= \frac{4\pi}{\int_{0}^{2\pi} \int_{0}^{\pi/2} \cos^{q}\theta \sin\theta \, d\theta \, d\varphi} 
		= 2q + 2,
	\end{equation}
	where $\Omega$ denotes the solid angle. Based on this, we have $D_B=2q_B+2$, $D_R=2q_R+2$, and $D_k=2q_k+2$, which are all measured in dB. Then, by substituting $D_B$, $D_R$, and $D_k$, equations (3) and (4) can be rewritten as ${\rho _{B,R}} = \frac{{{\rho _0}}}{{d_{B,R}^2}}{10^{0.2({q_B} + {q_R} + 2)}}{\cos ^{{q_B}}}\left( {\theta _{B,R}^{{\rm{AoD}}} - {\theta _{B0}}} \right){\cos ^{{q_R}}}\left( {\theta _{B,R}^{{\rm{AoA}}}} \right)$, and ${\rho _{R,k}} = \frac{{{\rho _0}}}{{d_{R,k}^2}}{10^{0.2({q_R} + {q_k} + 2)}}{\cos ^{{q_R}}}\left( {\theta _{R,k}^{{\rm{AoD}}}} \right){\cos ^{{q_k}}}\left( {\theta _{R,k}^{{\rm{AoA}}} - {\theta _{k0}}} \right)$, where $\theta_{B0}$ and $\theta_{R0}$ denote the antenna rotation angles of the BS and user $k$. It is evident that the optimal directions of the antennas are both pointing to the RIS/STAR-RIS, i.e. $\theta_{B0}=\theta_{B,R}^{\rm AoD}$ and $\theta_{k0}=\theta_{R,k}^{\rm AoA}$. Therefore, the path-loss  ${\rho _{B,R}}$ and ${\rho _{R,k}}$ are reformulated as
	\begin{equation}
		{\rho _{B,R}} = \frac{{{\rho _0}}}{{d_{B,R}^2}}{10^{0.2({q_B} + {q_R} + 2)}}{\cos ^{{q_R}}}\left( {\theta _{B,R}^{{\rm{AoA}}}} \right),
	\end{equation}
	\begin{equation}
		{\rho _{R,k}} = \frac{{{\rho _0}}}{{d_{R,k}^2}}{10^{0.2({q_R} + {q_k} + 2)}}{\cos ^{{q_R}}}\left( {\theta _{R,k}^{{\rm{AoD}}}} \right).
	\end{equation}
	
	To clearly characterize the location of the users and BS relative to the RIS/STAR-RIS, their elevation angles $\alpha_{k}$ and $\alpha_{{\rm BS}}$, as well as azimuth angles $\beta_{k}$ and $\beta_{{\rm BS}}$ with respective to the RIS/STAR-RIS are introduced. Accordingly, the unit direction vector from user $k$ and the BS to the RIS/STAR-RIS can be expressed as 
	\begin{equation}	 
		{{\bf{e}}_k} = {\left[ {\cos {\beta _k}\cos {\alpha _k},\cos {\beta _k}\sin {\alpha _k},\sin {\beta _k}} \right]^T},
	\end{equation}
	\begin{equation}	 
		{{\bf{e}}_{{\rm{BS}}}} = {\left[ {\cos {\beta _{{\rm{BS}}}}\cos {\alpha _{{\rm{BS}}}},\cos {\beta _{{\rm{BS}}}}\sin {\alpha _{{\rm{BS}}}},\sin {\beta _{{\rm{BS}}}}} \right]^T}.
	\end{equation}
	
	However, due to the different deployment orientations of the RIS and STAR-RIS, the calculation of $\theta_{B,R}^{\rm AoA}$ and $\theta_{R,k}^{\rm AoD}$ differs accordingly. For the RIS, we introduce its unit normal vector as ${\bf e_{\rm RIS }}=[0,0,-1]^T$, based on which $\theta^{\rm AoD}_{B,R}$ and $\theta^{\rm AoA}_{R,k}$ can be calculated by
	\begin{equation}	 
		\cos\theta _{B,R}^{{\rm{AoD}}} =  - {\bf{e}}_{{\rm{RIS}}}^T{{\bf{e}}_{{\rm{BS}}}} = \sin {\beta _{{\rm{BS}}}},
	\end{equation}
	\begin{equation}	 
		\cos\theta _{R,k}^{{\rm{AoA}}} =  - {\bf{e}}_{{\rm{RIS}}}^T{{\bf{e}}_k} = \sin {\beta _k}.
	\end{equation}
	
	In contrast, for the STAR-RIS, its vertical deployment makes its orientation critically important to system performance. To this end, we introduce an angle $\eta$ to represent its orientation. When $\eta = 0$, the normal vector of the STAR-RIS points along the positive $x$-axis, and when $\eta > 0$, it rotates counterclockwise. Therefore, its unit normal vector can be expressed as ${{\bf{e}}_{{\rm{STAR}}}} = {[\cos \eta ,\sin \eta ,0]^T}$. Based on this, $\theta^{\rm AoD}_{B,R}$ and $\theta^{\rm AoA}_{R,k}$ are given by
	\begin{equation}	 
			\cos \theta _{B,R}^{{\rm{AoD}}} =  - \cos {\beta _{{\rm{BS}}}}\left( {\cos {\alpha _{{\rm{BS}}}}\cos \eta  - \sin {\alpha _{{\rm{BS}}}}\sin \eta } \right),
	\end{equation}
	\begin{equation}	 
		\cos \theta _{R,k}^{{\rm{AoA}}} =  - \cos {\beta _k}\left( {\cos {\alpha _k}\cos \eta  - \sin {\alpha _k}\sin \eta } \right).
	\end{equation}
	It is worth noting that, unlike the aerial RIS whose $\theta _{R,k}^{{\rm{AoA}}} \in [0,\pi /2)$, the value of $\cos \theta _{R,k}^{{\rm{AoA}}}$ may become negative when the user is located in the transmission half-space of the STAR-RIS. Therefore, we take its absolute value during computation and determine the user type based on its sign, given by
	\begin{equation}	 
		k \in \left\{ {\begin{array}{*{20}{c}}
				{{{\cal K}_r},}&{\cos \theta _{R,k}^{{\rm{AoA}}} \ge 0,}\\
				{{{\cal K}_t},}&{\cos \theta _{R,k}^{{\rm{AoA}}} < 0.}
		\end{array}} \right.
	\end{equation}
	
	\subsection{Signal Model}
	In this paper, the direct BS-user channels are assumed to be blocked. Thus, in the aerial STAR-RIS-aided scenario, the received signal at user $k,\forall k \in {{\cal K}_i},\forall i \in \{ t,r\} $, is given by
	\begin{equation}	 
		{y_k} = \sum\limits_{j \in {\cal K}} {{{\bf h}_{R,k}^{H}}{{\bf \Theta} _i}{{\bf H}_{B,R}}{{\bf w}_j}{s_j}}  + {n_k},
	\end{equation}
	where ${{\bf{\Theta }}_i}  = {\rm{diag}}({\beta _{i,1}}{e^{j{\phi _{i,1}}}}, \ldots ,{\beta _{i,n}}{e^{j{\phi _{i,n}}}},\ldots,{\beta _{i,N}}{e^{j{\phi _{i,N}}}})= {\rm{diag}}\left( {{{\bm\theta} _i}} \right),i \in \{ t,r\}$ denotes the diagonal transmission and reflection coefficient matrix of the STAR-RIS. Here, $\beta_{i,n}$ and $\phi_{i,n}$ indicate the amplitude and phase shift of the $n$-th element in the transmission or reflection mode, respectively. ${{\bf{w}}_j} \in {\mathbb{C}^{M \times 1}}$ represents the active beamforming vector for user $j$ at the BS, while $n_k$ denotes the additive white Gaussian noise with power $\sigma^2$. Then, the achievable rate of user $k$ can be expressed as 
	\begin{equation}	 
		{R_k} = {\log _2}\left( {1 + \frac{{{{\left| {{\bf h}_{R,k}^H{{\bf\Theta} _i}{{\bf H}_{B,R}}{{\bf w}_k}} \right|}^2}}}{{\sum\nolimits_{j \in {\cal K}/k} {{{\left| {{\bf h}_{R,k}^H{{\bf\Theta} _i}{{\bf H}_{B,R}}{{\bf w}_j}} \right|}^2} + \sigma _k^2} }}} \right).
	\end{equation}

	Similarly, the achievable rate for user $k$ in the RIS-aided scenario can be derived accordingly.
	
	\subsection{Problem Formulation}
	In this paper, we aim to compare the performance of aerial RIS and STAR-RIS architectures by evaluating their sum rates under varying altitudes and deployment positions. The optimization problem for the STAR-RIS scenario is formulated as follows:
	\begin{subequations}
		\begin{align}
			&\quad \;\mathop {\max }\limits_{{\bf W},{{\bm \theta} _t},{{\bm \theta}_r}} \;\sum\limits_{k = 1}^K {{R_k}}\\
			&{\rm{s}}{\rm{.t}}{\rm{.}}\;{\rm{tr}}\left( {{{\bf W}^H}{\bf W}} \right) \le {P_{\max }},\\
			&\quad\;\;\,\beta _{t,n}^2 + \beta _{r,n}^2 = 1,\forall n \in {\cal N},\\
			&\quad\;\;\cos \left( {{\phi _{t,n}} - {\phi _{r,n}}} \right) = 0,\forall n \in {\cal N}.
		\end{align}
	\end{subequations}
	Constraint (18b) limits the total transmission power at the BS, where ${\bf W} = \left[ {{{\bf w}_1}, \ldots ,{{\bf w}_K}} \right]^T$ and $P_{\rm max}$ represents the maximum power budget. Constraint (18c) arises from the energy conservation principle, ensuring that the total power of transmission and reflection coefficients does not exceed unity. Constraint (18d) models the practical phase-shift coupling effect typically observed in STAR-RIS implementations.
	
	Since problem (18) is non-convex with respect to $\left\{ {{\bf W},{{\bm \theta} _t},{{\bm\theta} _r}} \right\}$, we reformulate the sum-rate maximization problem into an equivalent weighted mean square error minimization problem utilizing the WMMSE algorithm \cite{9133107}, given by
	\begin{subequations}
		\begin{align}
			&\mathop{\min}\limits_{{\bf{W}}, {\bm{\theta}}_t, {\bm{\theta}}_r, {\bm{\varpi}}, {\bm{\nu}}} \sum_{k=1}^K {\varpi_k e_k}\\
			&{\rm{s}}{\rm{.t}}{\rm{.}}\;18({\text{b}})-(18{\text{d}}),
		\end{align}
	\end{subequations}
	where ${\bm\varpi}  = {[{{\varpi} _1}, \ldots ,{\varpi _K}]^T}$ represents the weight vector, $e_k$ is the mean square error, calculated by
	\begin{equation}	 
		\begin{array}{l}
			{e_k} = {\left| {{\nu _k}} \right|^2}\left( {{{\sum\limits_{j \in {\cal K}} {\left| {{\bm\theta} _i^T{\rm{diag}}({\bf h}_{R,k}^H){{\bf H}_{B,R}}{{\bf w}_j}} \right|} }^2} + \sigma _k^2} \right)\\
			\quad \quad  - 2{\mathop{\rm Re}\nolimits} \left\{ {{\nu _k}{\bm\theta} _i^T{\rm{diag}}({\bf h}_{R,k}^H){{\bf H}_{B,R}}{{\bf w}_k}} \right\} + 1,
		\end{array}
	\end{equation}
	where ${\bm \nu}  = \left[ {{\nu _1}, \ldots ,{\nu _K}} \right]^T$ denote auxiliary variables. As demonstrated in \cite{9935266}, if $\{{\bf W},{{\bm\theta} _t},{{\bm\theta} _r}, {\bm \varpi} ,{\bm\nu}\} $ is an optimal solution to problem (19), $\{ {\bf W},{{\bm\theta} _t},{{\bm\theta} _r}\} $ is also an optimal solution to problem (18). 
	
	Notably, the objective function in problem (19) is block-wise convex corresponding to $\{ {\bm \varpi} ,{\bm\nu}\} $, $\bf W$, and  $\{ {{\bm\theta} _t},{{\bm\theta} _r}\} $, while the phase shift coupling constraint (18d) is non-convex. To address this issue, the penalty dual decomposition (PDD) method is adopted \cite{9120361}. Specifically, we introduce the auxiliary variables ${\tilde {\bm{\theta }_i}} = {[ {{\tilde \beta _{i,1}} {e^{j\tilde \varphi _{i,1}}}, \ldots ,{\tilde \beta _{i,N}} {e^{j\tilde \varphi _{i,N}}}} ]^T}$, ${\tilde {\bm{\beta }_i}} = {[ { {\tilde \beta _{i,1}} , \ldots , {\tilde \beta _{i,N}} }]^T}$, ${\tilde {\bm{\varphi }_i}} = {[ {{e^{j\tilde \varphi _{i,1}}}, \ldots ,{e^{j\tilde \varphi _{i,N}}}}]^T}$, and the constraint ${\tilde {\bm\theta} _i} = {{\bm\theta} _i},\quad i \in \{ r,t\}$. Furthermore, we incorporate the equation constraints on the auxiliary variables to the objective function as penalty terms to convert problem (19) to an augmented Lagrangian problem, given by
	\begin{subequations}
		\begin{align}
			&\mathop {\min }\limits_{{\bf{W}},{{\bm\theta} _t},{{\bm\theta} _r},{{\tilde {\bm\theta} }_t},{{\tilde {\bm\theta} }_r},\varpi ,\nu } \sum\limits_{k = 1}^K {{\varpi _k}{e_k}}  + \frac{1}{2}\sum\limits_{i \in \{ t,r\} } {{{\left\| {{{\tilde {\bm\theta} }_i} - {{\bm\theta} _i} + \rho {{\bm \lambda} _i}} \right\|}^2}}\\
			&\quad\quad\quad\quad\quad\;{\rm{s}}{\rm{.t}}{\rm{.}}\;\tilde \beta _{t,n}^2 + \tilde \beta _{r,n}^2 = 1,\forall n \in {\cal N},\\
			&\;\quad\quad\quad\quad\quad\quad\;\;\cos \left( {{{\tilde \phi }_{t,n}} - {{\tilde \phi }_{r,n}}} \right) = 0,\forall n \in {\cal N},\\
			&\;\quad\quad\quad\quad\quad\quad\;\;(18{\rm{b}}) - (18{\rm{d}}),
		\end{align}
	\end{subequations}
	where $\rho>0$ denotes the penalty factor, ${\bm\lambda}_t$ and ${\bm\lambda}_r$ are the Lagrangian dual variables. Then, the above problem can be solved through BCD by alternatively optimizing the blocks $\{ {\bm \varpi} ,{\bm\nu}\} $, $\bf W$, $\{ {{\bm\theta} _t},{{\bm\theta} _r}\} $, $\{\tilde{\boldsymbol{\beta}}_{t},\tilde{\boldsymbol{\beta}}_{r}\}$ and $\{\tilde{\boldsymbol{\phi}}_{t},\tilde{\boldsymbol{\phi}}_{r}\}$, respectively.
	
	\subsubsection{Update $\{ {\bm \varpi} ,{\bm\nu}\} $}
	The optimal $\varpi_k$ and $\nu_k$ are given by
	\begin{equation}	 
		{\varpi _k} = 1 + \frac{{{{\left| {{\bf{h}}_{R,k}^H{{\bf{\Theta }}_i}{{\bf{H}}_{B,R}}{{\bf{w}}_k}} \right|}^2}}}{{\sum\nolimits_{j \in K/k} {{{\left| {{\bf{h}}_{R,k}^H{{\bf{\Theta }}_i}{{\bf{H}}_{B,R}}{{\bf{w}}_j}} \right|}^2} + \sigma _k^2} }},
	\end{equation}
	\begin{equation}	 
		{\nu _k} = \frac{{{{\left| {{\bf{h}}_{R,k}^H{{\bf{\Theta }}_i}{{\bf{H}}_{B,R}}{{\bf{w}}_k}} \right|}^2}}}{{\sum\nolimits_{j \in K} {{{\left| {{\bf{h}}_{R,k}^H{{\bf{\Theta }}_i}{{\bf{H}}_{B,R}}{{\bf{w}}_j}} \right|}^2} + \sigma _k^2} }},
	\end{equation}
	\subsubsection{Update ${\bf W}$ and $\{ {\bm \theta _t},{\bm \theta _r}\}$}	
	Considering the objective function (21a) is convex with respect to ${\bf W}$ and $\{ {\bm \theta _t},{\bm \theta _r}\}$, the optimal solutions could be effectively obtained using existing optimization tools, such as CVX \cite{10475146}.
	\subsubsection{Update $\{\tilde{\boldsymbol{\beta}}_{t},\tilde{\boldsymbol{\beta}}_{r}\}$ and $\{\tilde{\boldsymbol{\phi}}_{t},\tilde{\boldsymbol{\phi}}_{r}\}$}
	Given the phase shift ${\bm\theta}_t$ and ${\bm\theta}_r$, the optimization problem with respective to  $\{\tilde{\boldsymbol{\beta}}_{t},\tilde{\boldsymbol{\beta}}_{r}\}$ and $\{\tilde{\boldsymbol{\phi}}_{t},\tilde{\boldsymbol{\phi}}_{t}\}$ is reformulated as
	\begin{subequations}
	\begin{align}
		&\min_{\tilde{\boldsymbol{\beta}}_{t}, \tilde{\boldsymbol{\phi}}_{t}, \tilde{\boldsymbol{\beta}}_{r}, \tilde{\boldsymbol{\phi}}_{r}}  \sum_{i \in \{t,r\}} \mathrm{Re}\left(\boldsymbol{\vartheta}_{i} \mathrm{diag}(\tilde{\boldsymbol{\beta}}_{i}) \tilde{\boldsymbol{\phi}}_{i}\right) \\
		&\quad\quad\;\quad{\rm{s}}{\rm{.t}}{\rm{.}}\; (21{\rm{b}}),(21{\rm{c}}),
	\end{align}
	\end{subequations}
	where $\boldsymbol{\vartheta}_{i}=-{\bm\theta}_i+\rho{\bm\lambda_i},\quad i \in \{ r,t\}$ is the constant term. Although the problem remains non-convex, the closed-form solutions for $\tilde{\boldsymbol{\beta}}_{i}$ and $\tilde{\boldsymbol{\phi}}_{i}$ can be derived \cite{9935266}, respectively.
	
	For given $\tilde{\boldsymbol{\phi}}_{t}$ and $\tilde{\boldsymbol{\phi}}_{t}$, by defining ${{\boldsymbol\varphi} _i} = {\rm diag}(\tilde {\boldsymbol\phi} _i^H){{\boldsymbol\vartheta} _i}$, ${p_n} = \left| {{{\boldsymbol\varphi} _t}(n)} \right|\cos (\angle {{\boldsymbol\varphi} _t}(n))$, ${q_n} = \left| {{{\boldsymbol\varphi} _r}(n)} \right|\sin (\angle {{\boldsymbol\varphi} _r}(n))$, ${\xi _n} = {\mathop{\rm sgn}} ({q_n})\arccos (\frac{{{q_n}}}{{\sqrt {p_n^2 + q_n^2} }})$, the optimal $\tilde{\boldsymbol{\beta}}_{t}$ and $\tilde{\boldsymbol{\beta}}_{r}$ can be calculated by
	\begin{equation}	 
		{\tilde \beta _{t,n}} = \sin {\psi _n},\;\;{\tilde \beta _{r,n}} = \cos {\psi _n},
	\end{equation}
	where
	\begin{equation}	 
		{\psi _n} = \left\{ {\begin{array}{*{20}{ll}}
				{ - \frac{\pi }{2} - {\xi _n},}&{{\rm{if}}\;{\xi _n} \in [ - \pi , - \frac{\pi }{2}),}\\
				{0,}&{{\rm{if}}\;{\xi _n} \in [ - \frac{\pi }{2},\frac{\pi }{4}),}\\
				{\frac{\pi }{2},}&{\text{otherwise}.}
		\end{array}} \right.
	\end{equation}

	For fixed $\tilde{\boldsymbol{\beta}}_{t}$ and $\tilde{\boldsymbol{\beta}}_{r}$, let ${\tilde {\bm\varphi} _i} = {\text{diag}}(\tilde{ \boldsymbol{\beta}} _i^H){\tilde {\bm{\vartheta}} _i}$, $\omega _n^ +  = {{\tilde {\bm\varphi} }_t}(n) + j{{\tilde {\bm\varphi} }_r}(n)$, $\omega _n^ -  = {{\tilde {\bm\varphi} }_t}(n) - j{{\tilde {\bm\varphi} }_r}(n)$, the closed-form solutions for $\tilde{\boldsymbol{\phi}}_{t}$ and $\tilde{\boldsymbol{\phi}}_{t}$ are given by
	\begin{equation}	 
		\left( {{{\tilde \phi }_{t,n}},{{\tilde \phi }_{r,n}}} \right) = \mathop {\arg \min }\limits_{\left( {{{\tilde \phi }_{t,n}},{{\tilde \phi }_{r,n}}} \right) \in \chi _\phi ^n} {\mathop{\rm Re}\nolimits} \left( {{{\tilde \vartheta }_{t,n}}{{\tilde \phi }_{t,n}}} \right) + {\mathop{\rm Re}\nolimits} \left( {{{\tilde \vartheta }_{r,n}}{{\tilde \phi }_{r,n}}} \right),
	\end{equation}
	where
    \begin{equation}	 
    \scalebox{0.95}{$
		\chi _\phi ^n = \left\{ {\left( {{e^{j(\pi  - \angle \omega _n^ + )}},{e^{j(\frac{3}{2}\pi  - \angle \omega _n^ + )}}} \right),\left( {{e^{j(\pi  - \angle \omega _n^ - )}},{e^{j(\frac{1}{2}\pi  - \angle \omega _n^ - )}}} \right)} \right\}.
        $}
	\end{equation}
	
	Therefore, the dual variables $\{{\bm \lambda}_t,{\bm \lambda}_r\}$ and penalty factor $\rho$ can be updated based on PDD framework. 

	\section{simulation results}
	In this section, simulation results are presented to evaluate the performance of aerial RIS and STAR-RIS architectures under varying deployment positions, altitudes, and orientations.
	
	In the simulation setup, the BS is located at $(0, 0, 0)$ $\mathrm{m}$ with the number of antennas $M = 8$ and the maximum transmission power $P_{\max} = 20\, \text{dB}$. There are $K = 4$ users, which are randomly distributed within a $100\times100\text{ }\mathrm{m}^2$ square region defined by the coordinates $(0, 0, 0)$ $\mathrm{m}$ and $(100, 100, 0)$ $\mathrm{m}$. The directivity parameters are $q_B = 20$ for the BS, $q_k = 20$ for the users, and $q_R = 3$ for both the RIS and STAR-RIS. The path-loss factor is $\rho_0 = 1\, \text{dBm}$, and the noise power is $\sigma^2 = -70\, \text{dBm}$.
	\begin{figure}[t]
		\centering
		\includegraphics[width=1\columnwidth]{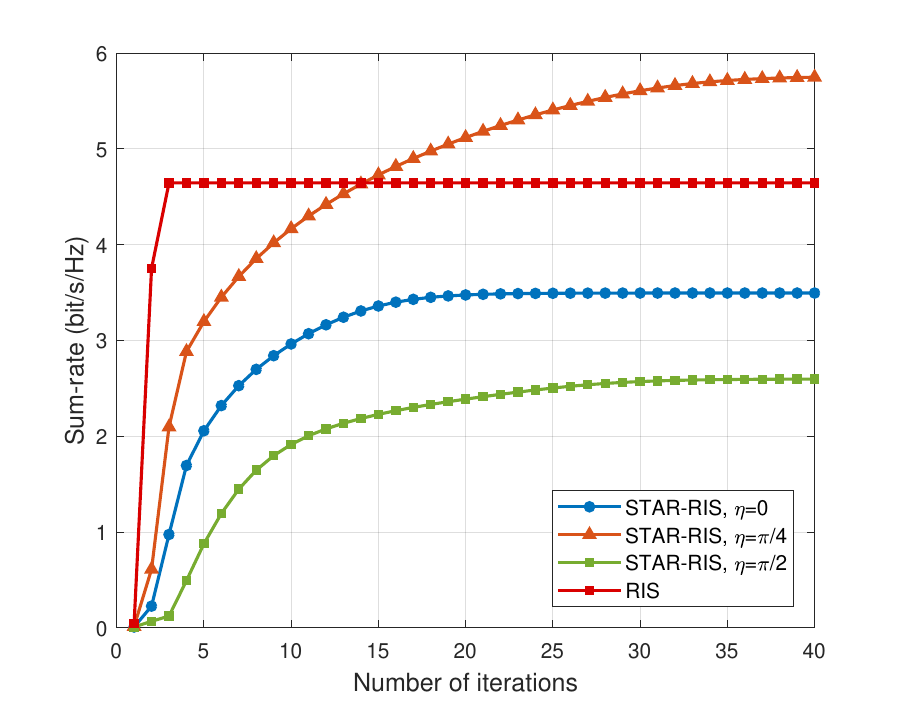}
		\caption{The convergence performance of the proposed algorithm.} 
	\label{fig:sys-model}
	\end{figure}
	\begin{figure}[t]
		\centering
		\begin{subfigure}[htbp]{0.24\textwidth}
			\centerline{\includegraphics[width=\textwidth]{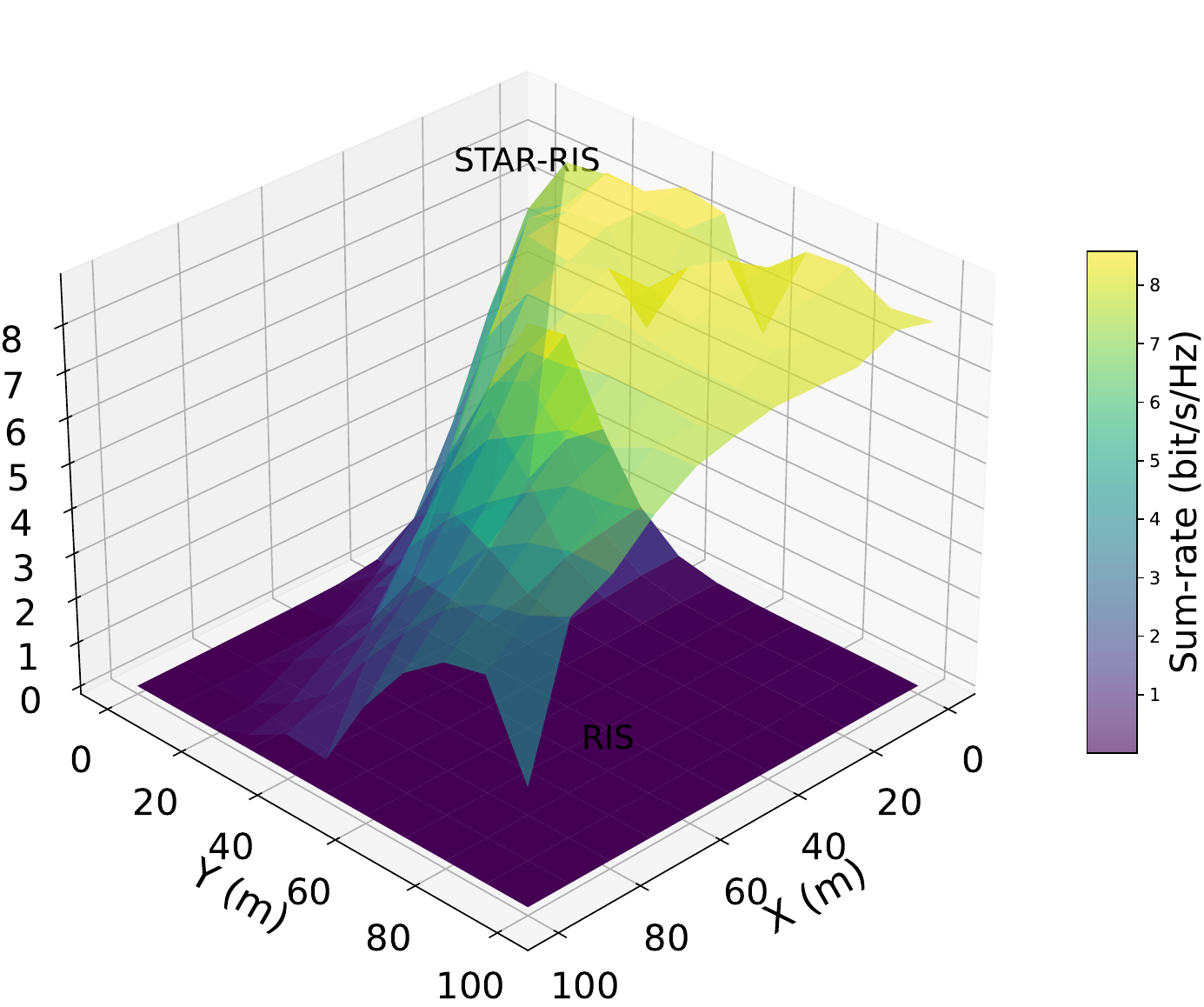}}
			\caption{$H=10$ m.}
			\label{fig:tra1}
		\end{subfigure}
		\begin{subfigure}[htbp]{0.24\textwidth}
			\centerline{\includegraphics[width=\textwidth]{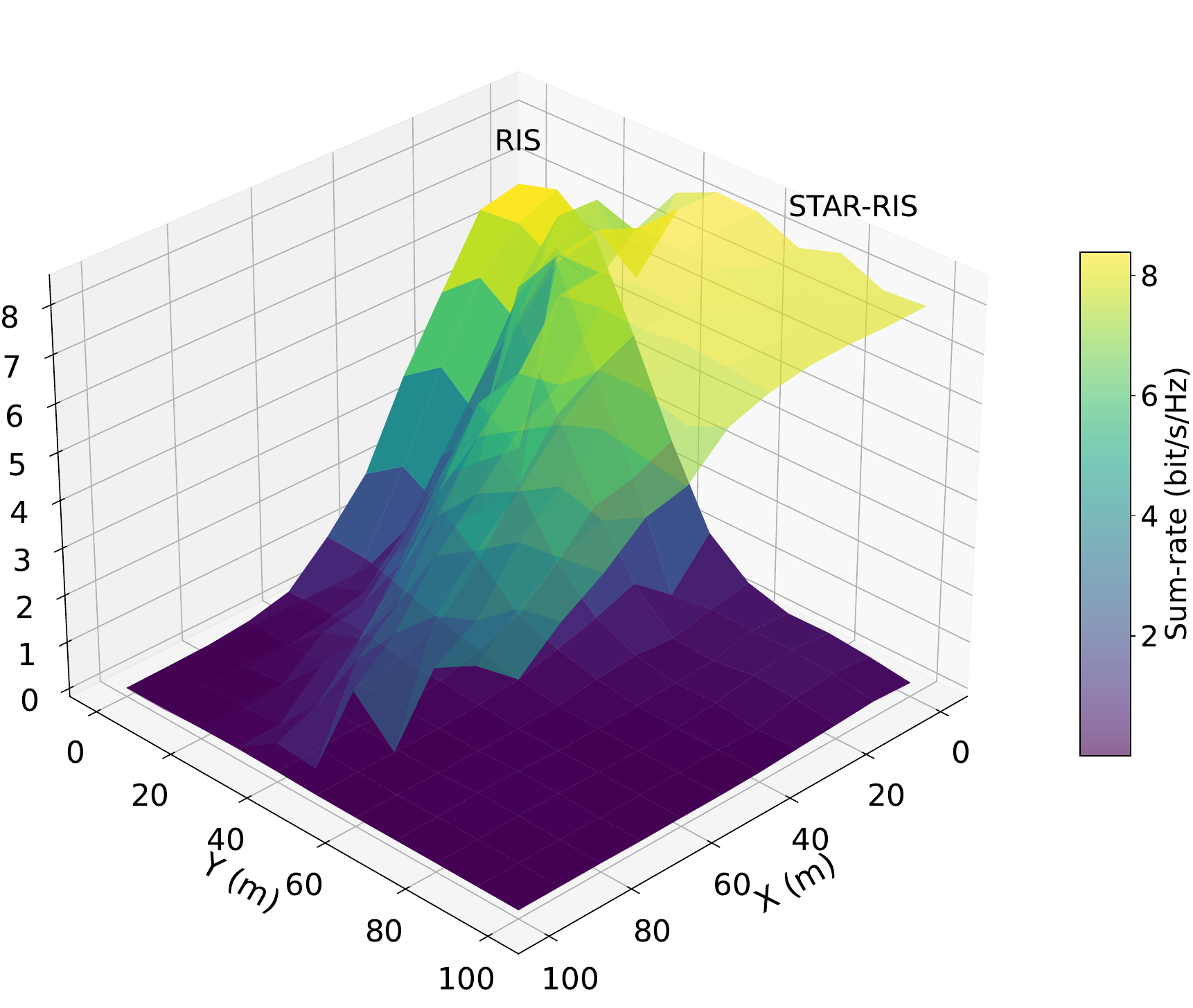}}
			\caption{$H=20$ m.}
			\label{fig:tra2}
		\end{subfigure}
		\begin{subfigure}[htbp]{0.24\textwidth}
			\centerline{\includegraphics[width=\textwidth]{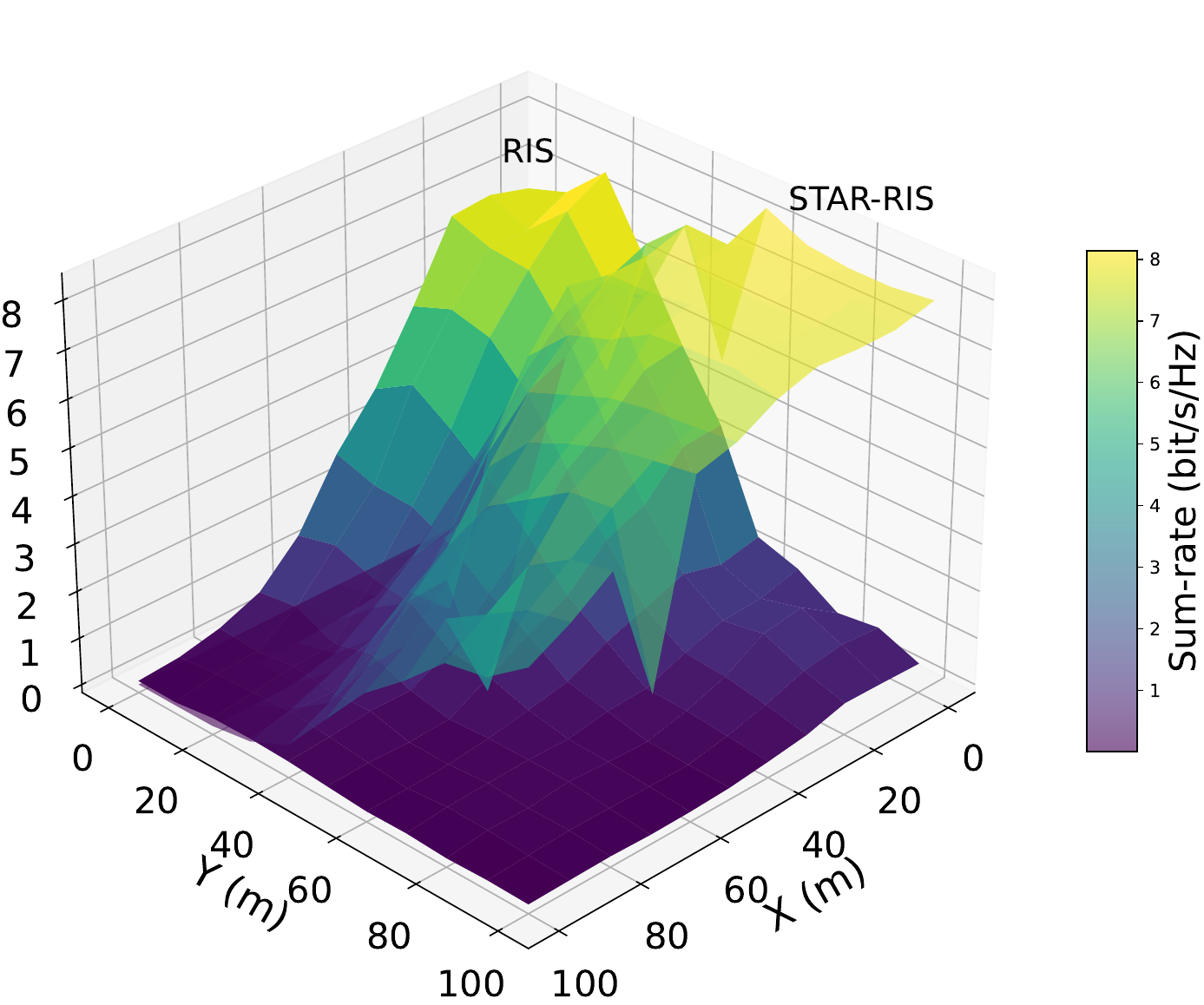}}
			\caption{$H=30$ m.}
			\label{fig:tra3}
		\end{subfigure}
		\begin{subfigure}[htbp]{0.24\textwidth}
			\centerline{\includegraphics[width=\textwidth]{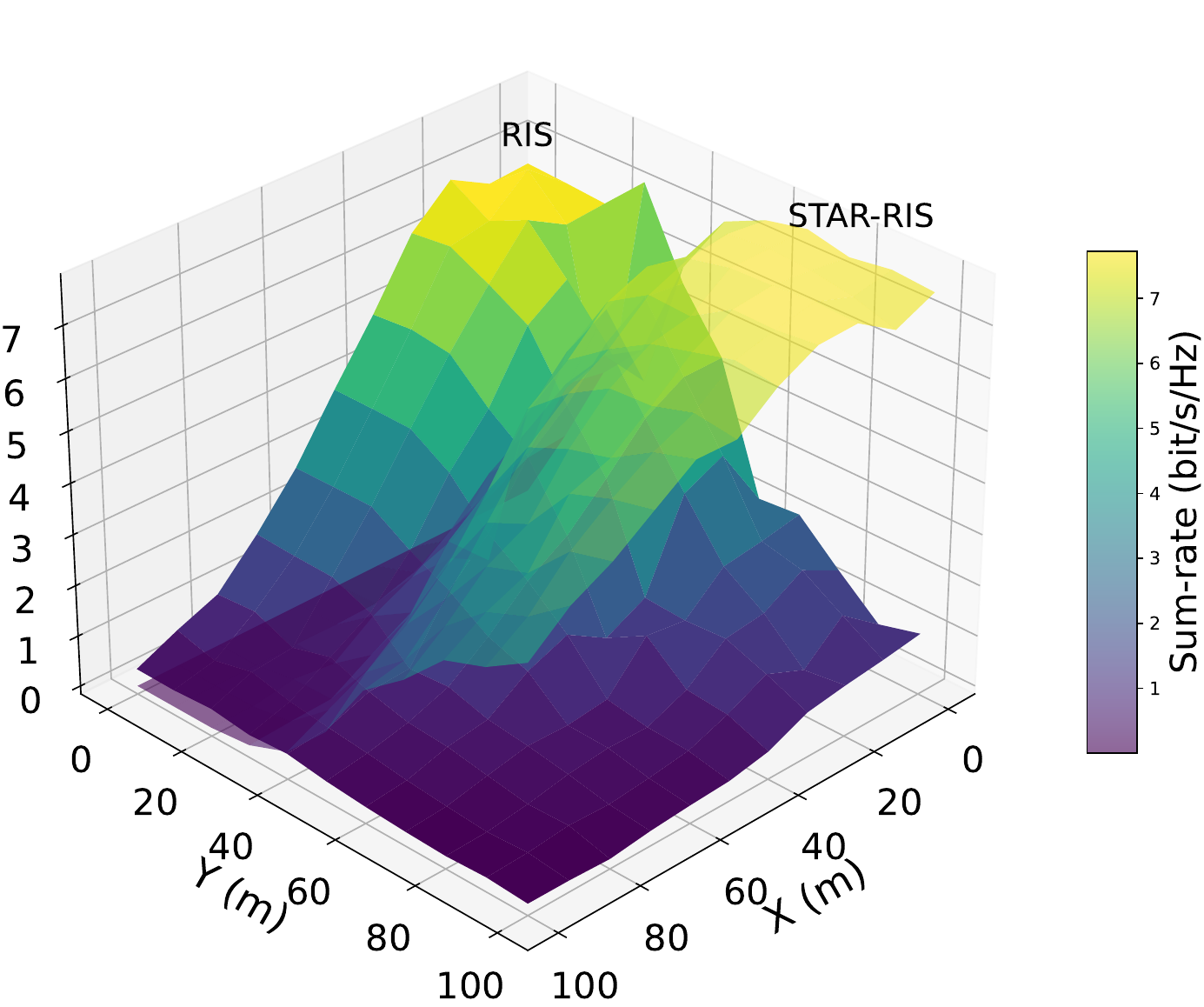}}
			\caption{$H=40$ m.}
			\label{fig:tra4}
		\end{subfigure}
		\caption{Effect of various deployment positions and altitudes on the sum rate of the RIS and STAR-RIS architectures.}
	\end{figure}

	\begin{figure}[t]
		\centering
		\begin{subfigure}[htbp]{0.24\textwidth}
			\centerline{\includegraphics[width=\textwidth]{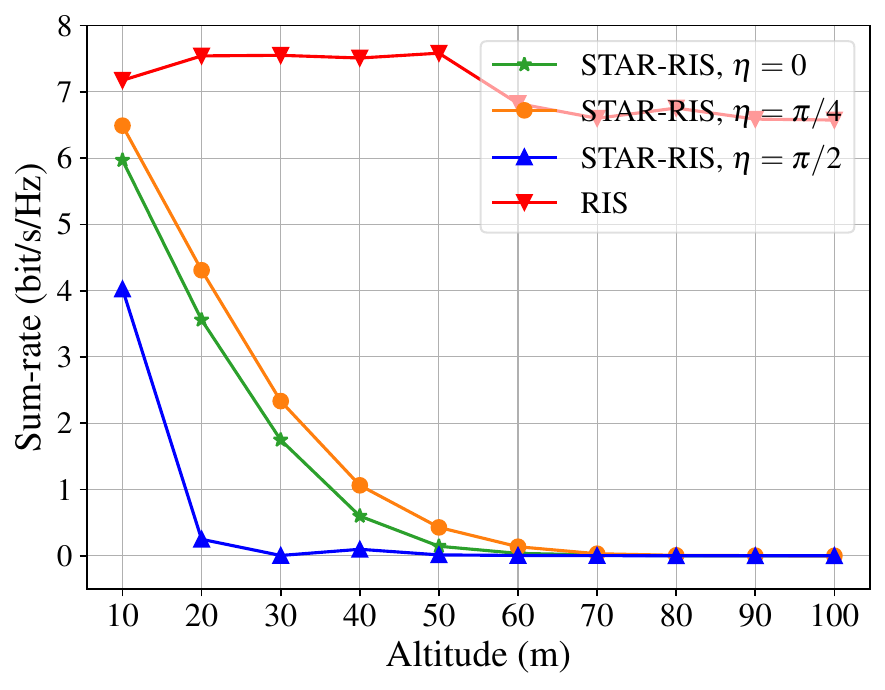}}
			\caption{$(x,y)=(10,10)$ m.}
			\label{fig:tra1}
		\end{subfigure}
		\begin{subfigure}[htbp]{0.24\textwidth}
			\centerline{\includegraphics[width=\textwidth]{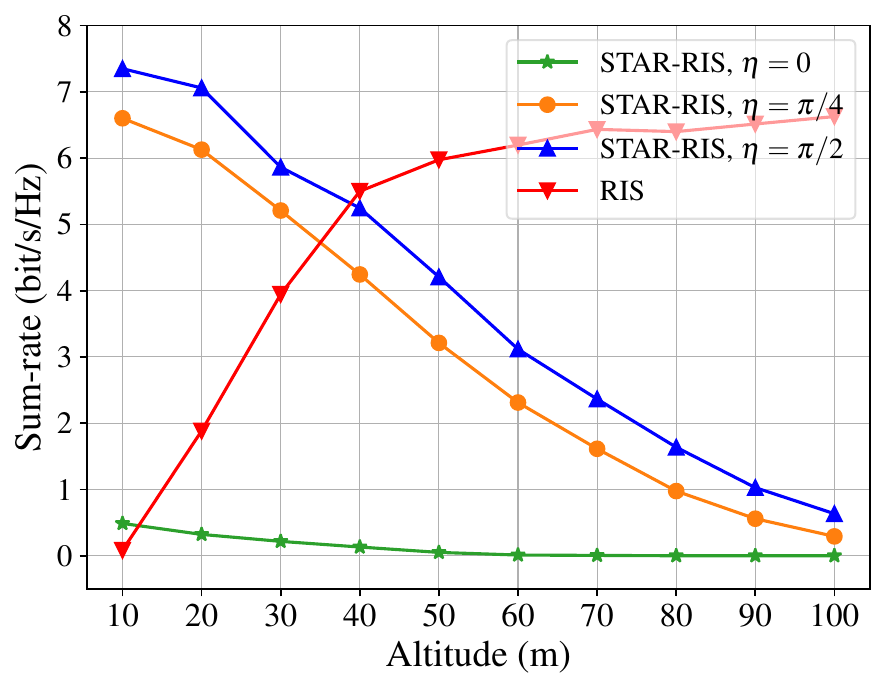}}
			\caption{$(x,y)=(20,60)$ m.}
			\label{fig:tra2}
		\end{subfigure}
		\begin{subfigure}[htbp]{0.24\textwidth}
			\centerline{\includegraphics[width=\textwidth]{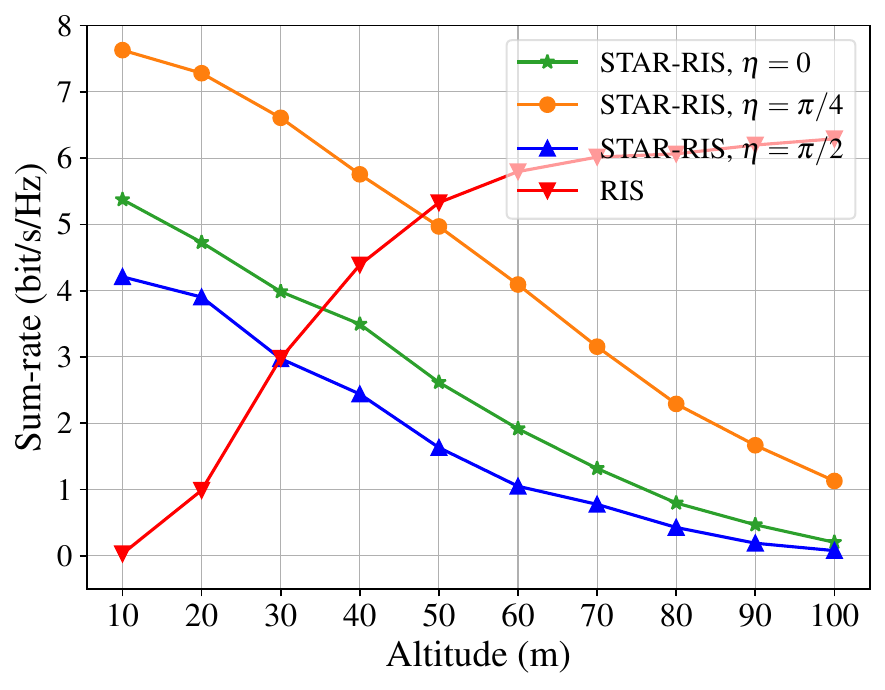}}
			\caption{$(x,y)=(50,50)$ m.}
			\label{fig:tra3}
		\end{subfigure}
		\begin{subfigure}[htbp]{0.24\textwidth}
			\centerline{\includegraphics[width=\textwidth]{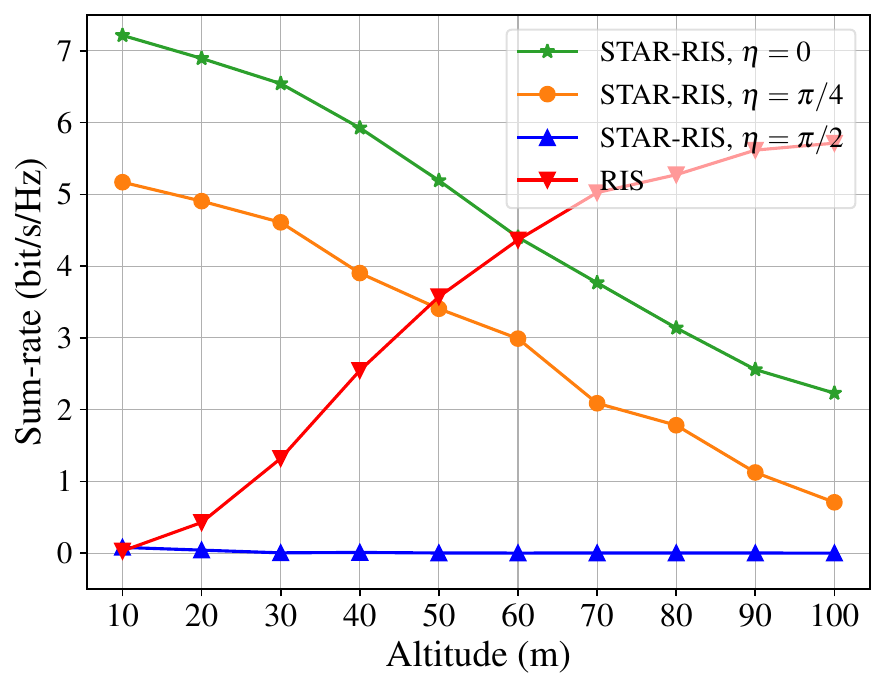}}
			\caption{$(x,y)=(80,20)$ m.}
			\label{fig:tra4}
		\end{subfigure}
		\caption{Effect of altitude and orientation on the sum rate of the RIS and STAR-RIS architectures.}
	\end{figure}

	In Fig. 2, we compare the convergence performance of the proposed algorithm under different orientations, where the number of elements is set to $N=20$ and the STAR-RIS/RIS is deployed at the position of $(50,50,40)\,\mathrm{m}$. It can be clearly observed that the proposed algorithm rapidly converges to a stable value under all orientations. Moreover, the performance of the STAR-RIS is highly sensitive to its orientation, with the optimal performance obtained when the angle $\eta = \pi/4$.
	
	In Fig. 3, we compare the sum-rate performance of the aerial RIS and STAR-RIS architectures under varying deployment positions and altitudes, with the orientation of the STAR-RIS set to $\eta = 0$. It is evident that at a lower altitude, the STAR-RIS significantly outperforms the RIS. This performance advantage is attributed to the full-space coverage capability of the STAR-RIS, which enables efficient signal propagation in both the transmission and reflection regions. In contrast, the RIS, deployed horizontally, suffers from large incident and reflection angles at low altitudes, resulting in reduced effective channel gain. As the altitude $H$ increases, the performance of both architectures changes noticeably. Specifically, the RIS gradually exhibits better performance in regions closer to the base station due to improved angular alignment, while the STAR-RIS consistently maintains superior performance in areas farther away, owing to its broader angular adaptability.

    To further illustrate the impact of altitude and orientation on aerial RIS and STAR-RIS, we compare the performance of the two architectures at different altitudes under four horizontal positions in Fig. 4. Aerial RIS demonstrates superior performance at higher altitudes when deployed close to the BS. Due to its horizontal orientation, RIS benefits from better signal alignment with the base station at elevated positions, resulting in enhanced signal propagation and higher sum-rate. On the other hand, aerial STAR-RIS excels at lower altitudes when situated farther from the base station. Thanks to its vertical deployment, STAR-RIS can provide full-space coverage, simultaneously transmitting and reflecting signals, ensuring reliable performance even at low altitudes or at greater distances from the BS. This full-space capability allows STAR-RIS to maintain high communication rates in scenarios where RIS faces limitations due to suboptimal angle alignments. However, the performance of STAR-RIS is highly sensitive to its orientation. For example, in Fig. 4 (d), when the deployment angle is set to $\eta=\pi/2$, the performance of aerial STAR-RIS significantly deteriorates, which demonstrates that the orientation is important for the deployment of aerial STAR-RIS.

\balance
	
	\section{conclusion}
	This paper conducted a systematic performance comparison between aerial RIS and STAR-RIS architectures under various deployment positions, altitudes, and orientation configurations. By formulating the sum-rate maximization problem and developing an efficient algorithm based on WMMSE and BCD methods, we thoroughly investigated the spatial characteristics and performance trends of both architectures. Simulation results reveal that STAR-RIS achieves superior performance in low-altitude scenarios due to its ability to provide full-space coverage, whereas conventional RIS demonstrates better performance in high-altitude deployments near the base station. These insights contribute to a better understanding of RIS and STAR-RIS deployment strategies and offer practical guidance for the design of future 6G communication networks.
	
	\bibliographystyle{IEEEtran}
	\bibliography{references.bib}  
\end{document}